# Tracking System of Coal Mine Patrol Robot for Low-Illumination Environments

Shaoze You[1,2], Hua Zhu[1,2,*], Menggang Li[1,2], Yutan Li[1,2,3]

[1] School of Mechanical and Electrical Engineering, China University of Mining and Technology, Xuzhou 221116, China

[2] Jiangsu Collaborative Innovation Center of Intelligent Mining Equipment, China University of Mining and Technology, Xuzhou 221008, China

[3] School of Intelligent Manufacturing, Jiangsu Vocational Institute of Architectural Technology, Xuzhou 221116, China

zhuhua83591917@163.com

**Abstract:** Computer vision, an integral component for robots to apperceive the external environment, has received a significant amount of attention in recent years. Discriminative Correlation Filter (DCF)-based trackers have gained more popularity due to their efficiency; however, tracking in low-illumination environments remains a challenging problem that has not yet been successfully addressed in the existing literature. In this work, this problem is addressed via the introduction of a Low-Illumination Long-term Correlation Tracker (LLCT). First, fused features only including HOG and Color Names are employed to boost the tracking efficiency. Second, the standard PCA is used to as the reduction scheme in the translation and scale estimation phase for acceleration. Third, a long-term correlation filter is learned to keep maintain the long-term memory ability of the tracker. Finally, the memory templates are updated with interval updates, and then re-match the existing and initial templates are re-matched every few frames to maintain template accuracy. Extensive experiments on the popular Object Tracking Benchmark OTB-50 datasets have demonstrated that the proposed tracker significantly outperforms the state-of-the-art trackers, and significantly achieves a real-time (33 FPS) performance. In addition, the proposed tracker can be applied to coal mine patrol robots working in low-illuminance illumination environments. The experimental results show that the novel tracker performance in low-illumination environment is better than that of general trackers.

## 1 Introduction

Visual object tracking (VOT) is one of the most important fundamental problems in computer vision, and also plays a central role in real-time vision applications, such as intelligent monitoring system, automatic driving, and robotics [1-3]. The objective is to select the target from the first frame and track it in the subsequent frame by inputting a video frame or real-time image.

In recent years, discriminative correlation filter (DCF)-based methods, including MOSSE [4], KCF [5], CSR-DCF [6], BACF [7] and ECO [8], have significantly advanced the state-of-the-art technology for short-term tracking. Due to the online nature of tracking and its requirements for practical applications, an ideal tracker should be accurate and robust for a longer period of time and in real-time vision systems. Moreover, due to the complexity of the working environment, including appearance deformation object occlusion, illumination change, motion blur, and objects out of view, the performance of trackers is limited, which can cause the tracker to drift easily during long-term tracking.

In recent years, mobile robots have become a research hotspot in science and technology. The path planning, positioning and navigation, obstacle avoidance and other aspects of mobile robot are inseparable from the assistance of vision [9-11]. Because the mobile robots are limited by their endurance, volume, and flexibility, the hardware configuration of the robot is usually based on low power consumption, thus, an algorithm that requires a large amount of computation can seriously affect the real-time performance of the robot. In practical engineering applications, because of the blindness effect of cameras can be caused by low-illumination conditions due to light mutation. This phenomenon is particularly serious in coal mines, in which the light distribution interval is large, and affects the performance of the visual tracker.

Hence, this work proposes the development of a long-term real-time correlation filter (LRCF) tracker for long-term object tracking in low-illumination environments. The proposed method can learn/update filters from real negative examples that are densely extracted from the background. The SVM classifier is used as a heavy detector to achieve the possibility of long-term tracking, and has better tracking accuracy and real-time performance. Finally, to adapt to the working environment of coal mine robots, the image exposure compensation for low-illumination environment is added, and the Low-Illumination Long-term Correlation Tracker (LLCT) is proposed. The flow chart of the proposed method is illustrated in Fig. 1.

The paper is organized as follows. Section 2 discusses previous work related to the proposed tracking framework. The classical DCF tracking formulation and our approach are introduced in Section 3. The experimental results are analyzed and discussed in Section 4. Finally, the conclusion is presented in Section 5.

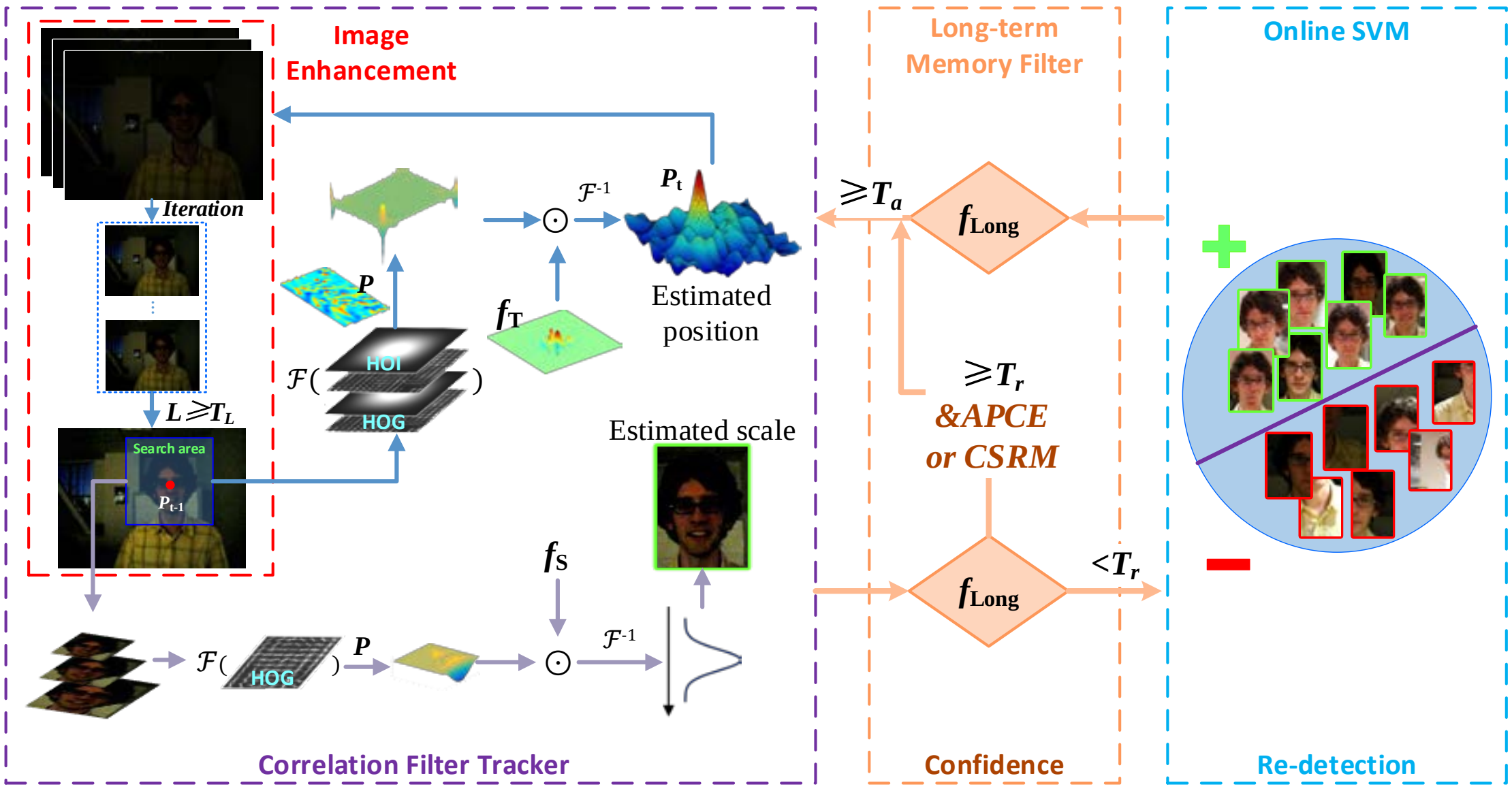

**Fig. 1.** The flow chart of the proposed method. The projection matrix $P$ was applied to the LCT [18] architecture to reduce the amount of computation.

## 2 Related Work

In this section, the recent achievements related to the present work are described. For a comprehensive overview of existing tracking methods, readers can refer to the following surveys and evaluations.

**Tracking-by-detection.** The tracking-by-detection method regards the target tracking in each frame as a detection problem in a local search window, and usually separates the target from its surrounding background by an incremental learning classifier [5, 12-14]. In this work, every frame is regarded as a single target detection process, which is the most commonly used method in visual object tracking.

**Correlation Filters tracker.** In 2011, Bolme et al. learned the MOSSE [4] filter for tracking with an impressive speed of more than 600 frames per second (FPS), and demonstrated the potential of the correlation filter. Discriminant correlation filter-based (DCF-based) trackers have emerged in endlessly, and numerous improved DCF-based trackers [4-8, 13, 15] that exhibit more precise and robust tracking performance by sacrificing the tracking speed have been proposed. By using the properties of fast Fourier transform (FFT) and inverse fast Fourier transform (IFFT), the DCF-based tracker greatly reduces the computing time by means of correlation operation on image features.

**CNN-based tracker.** In recent years, deep learning (DL) has been widely considered. As a representative architecture, convolutional neural network (CNN) has achieved remarkable results in visual tracking due to its powerful feature expression ability [14, 16, 17]. It can attain a surprising high-precision result with the support of the graphics card, but its large amount of required computation has limited its development on mobile devices.

**Long-term tracker.** Long-term tracking is different from short-term tracking, as it has long-term occlusion and field-of-view situations. To achieve long-term tracking, Zdenek et al. proposed the tracking framework TLD [18] and decomposed the long-term tracking task into three parts: tracking, learning, and detection. After the correlation filter-based tracker became a research hotspot, Ma et al. proposed LCT [19] and LCT+ [20] which added re-detector and long-time memory templates to the short-term tracker. Zhu et al. proposed a novel collaborative correlation tracker (CCT) [21] using multi-scale kernelized the correlation Tracking (MKC) and online CUR filter for long-term tracking.

## 3 The Proposed Approach

### 3.1 Discriminative Correlation Filter

In this section, we adopt the accelerated version fast-DSST of discriminative scale space tracker (DSST) proposed by Danelljan et al. [22] is adopted as the baseline due to its outstanding performance. It learned two optional correlation filters for translation locating (2-dimensional) and estimation scale (1-dimensional) of the target in the new frame. For a 2-dimensional ($\mathrm{M} \times \mathrm{N}$) image, the goal is to learn a set of multichannel correlation filters $\mathbf{f}_t \in \mathbb{R}^{\mathrm{M}\times\mathrm{N}\times\mathrm{D}}$ based on the sample $\{(\mathbf{x}_t, y_t)\}_t^l$. Each training sample $\mathbf{x}_t \in \mathbb{R}^{\mathrm{M}\times\mathrm{N}\times\mathrm{D}}$ contains D-dimensional features extracted from the interest region, and the $l \in \{1, \dots, \mathrm{D}\}$ of $\mathbf{x}_t$ is denoted as $\mathbf{x}_t^l$. The correlation response label of the filter is expressed by $\mathbf{y}_t \in \mathbb{R}^{\mathrm{M}\times\mathrm{N}}$. This achieved by minimizing the $\ell_2$ error of the correlation response of the desired correlation filter $\mathbf{f}^l$.

$$\epsilon(\mathbf{f}) = \left\| \sum_{l=1}^{D} \mathbf{x}_t^l \circledast \mathbf{f}^l - \mathbf{y}_t \right\|_2^2 + \lambda \sum_{l=1}^{D} \left\| \mathbf{f}^l \right\|^2, \tag{1}$$

where $\circledast$ denotes circular convolution and $\lambda$ is a regularization weight. The response label $y$ is usually expressed by Gaussian function [4].

Because Eq. (1) is a linear least squares problem, it can be computed efficiently in Fourier domain by Parseval's formula. Hence, the filter that minimizes Eq. (1) is given by:

$$\epsilon(\mathbf{F}) = \left\| \sum_{l=1}^{D} \bar{\mathbf{X}}^l \odot \mathbf{F}^l - \mathbf{Y} \right\|_2^2 + \lambda \sum_{l=1}^{D} \left\| \mathbf{F}^l \right\|^2, \tag{2}$$

In Eq. (2), the capital letters denote the discrete Fourier transform (DFT) of the corresponding quantities, the bar $\bar{\bullet}$ denotes complex conjugation, and $\odot$ denotes a Hadamard product. Therefore, in the first frame, Eq. (2) can be solved as:

$$\mathbf{F}^l = \frac{\bar{\mathbf{Y}} \odot \mathbf{X}^l}{\sum_{k=1}^{D} \bar{\mathbf{X}}^k \odot \mathbf{X}^k + \lambda}, \qquad l = 1, \dots, D. \tag{3}$$

The numerator $A_t^l$ and the denominator $B_t$ are defined for the $t$-th frame. An optimal update strategy for the filter $\mathbf{F}_t^l$ to the new sample $x_t$ is as follows:

$$A_t^l = (1-\eta) A_{t-1}^l + \eta \bar{\mathbf{Y}} \odot \mathbf{X}_t^l, \tag{4}$$

$$B_t = (1-\eta) B_{t-1} + \eta \sum_{k=1}^{D} \bar{\mathbf{X}}_t^k \odot \mathbf{X}_t^k, \tag{5}$$

where the scalar $\eta$ is a parameter of the learning rate. To detect the variations of a position in a new frame $t$, the correlation scores $\mathbf{y}_t$ for a new test sample $\mathbf{z}_t$ can be computed in the Fourier domain

$$y_t = \mathcal{F}^{-1} \left\{ \frac{\sum_{l=1}^{D} \bar{A}_{t-1}^l \odot \mathbf{Z}^l}{B_{t-1} + \lambda} \right\}, \tag{6}$$

where $\mathbf{Z}^l$ denotes the $l$-dimensional features extracted from the frame of pending detection. $\mathcal{F}^{-1}$ is the IFFT. Eq. (6) can be used to find the maximum correlation score to determine the position of the current target.

To estimate the scale, the image feature pyramid of the current sample is constructed in a rectangular area behind learning translation filter, and define the scale filter as $S = \{\alpha^n | \lfloor -\frac{N-1}{2} \rfloor, \dots, \lfloor \frac{N-1}{2} \rfloor\}$. The current target region with a size of $W \times H$ is reconstructed to form a series of scale patches $I_n$ of size $\alpha^n W \times \alpha^n H$ based on $N$ scale levels. The scale filter has a 1-dimensional Gaussian score $y_s$. The max value $S_t(n)$ of the training sample $x_{t,scale}$ for $I_n$ is the current scale.

Because the computational cost of DSST is mainly determined by FFT, we use the method of PCA dimensionality reduction to improve the computing speed. To update the target template $\mu_t =$ $(1-\eta)\mu_{t-1}+\eta x_t$, a projection matrix $\mathbf{P}_t \in \mathbb{R}^{d\times D}$ is constructs for $\mu_t$, where *d* is the dimension of the compression feature. The current test sample $z_t$ can be obtained by Eq. (7) via the compressed training sample $\mathcal{X}_t = \mathcal{F}\{\mathbf{P}_{t-1}x_t\}$:

$$y_t = \mathcal{F}^{-1}\left\{\frac{\sum_{l=1}^{d}\bar{\mathcal{A}}_{t-1}^{l}\odot\mathcal{Z}_t^{l}}{\mathcal{B}_{t-1}+\lambda}\right\}, \tag{7}$$

where $\mathcal{Z}_t = \mathcal{F}\{\mathbf{P}_{t-1}\mathbf{z}_t\}$ is the new compressed sample, $\bar{\mathcal{A}}_{t-1}^{l}$ and $\mathcal{B}_{t-1}$ are the updated numerator and denominator of the template after feature compression, respectively. Note that the projection matrix $\mathbf{P}_t$ is not calculated explicitly, and can quickly be obtained by the QR-decomposition.

### 3.2 Long-Term Memory

In practical applications, trackers often work for a long time, and objects out of view and occlusion have been the main problems in long-term tracking; because when the target disappears, the reappearance of the position is often not the vanishing position. In special environments such as coal mines, light sources are dim and often not evenly distributed. When entering the dark light scene from the light environment, the camera has a blind eye effect due to the sudden exposure of the light source, which is also an important problem to be solved in long-time tracking. Therefore, the tracker needs to re-detect the location of the target, and also needs to store memories of the historical appearance of the target to prevent the trace from failing.

To avoid the tracking failure caused by the noise pollution of the model during the long-term tracking, we have been inspired by the work of LCT [18, 19], learned a long-term filter $\mathbf{f}_{Long}$. Unlike the LCT [18], we do not use the kernel trick [13] for $\mathbf{f}_{Long}$ to calculate the response score; instead, a DSST-like approach to directly calculate the correlation response between the test sample $\mathbf{x}_{t,Long}$ and the filter $\mathbf{f}_{Long}$. Note that the test sample $\mathbf{x}_{t,Long}$ for long-term filter $\mathbf{f}_{Long}$ can also use projection matrix $\mathbf{P}_t$ for feature compression, and in this work, the sample obtained from the translation filter is used as the detection sample of the long-term filter. For each tracked target $\mathcal{Z}_t = \mathcal{F}\{\mathbf{P}_{t-1}\mathbf{z}_t\}$, its confidence score is computed as $C_{t,Long} = \max\left(\mathbf{f}_{Long}(\mathcal{Z}_t)\right)$.

### 3.3 Re-detection Module

The re-detection module is an important part of improving the robustness and long-term tracking ability of the tracker. It is used to find the target quickly after the target is lost. Our method is built upon the LCT+ [19] method, and online SVM classifier is used as the detector. The difference is that another confidence parameter (Section 3.3), together with the predefined re-detection threshold $T_r$, is used as the criterion. The re-detector only trains the translated samples to further reduce the computational burden. The feature representation of the sample is based on the method proposed by MEEM [23], namely the quantized color histogram. For a training set $\{(\mathbf{v}_i,\ c_i)|i = 1,2,\dots,N\}$ with $N$ samples in a frame, the objective function of solving the SVM detector hyperplane $\mathbf{h}$ is:

$$\min_{\boldsymbol{h}} \frac{\lambda}{2}\|\mathbf{h}\|^2 + \frac{1}{N}\sum_i \ell\big(\mathbf{h}; (\mathbf{v}_i, c_i)\big),$$
$$\text{where } \ell\big(\mathbf{h}; (\mathbf{v}, c)\big) = \max\{0, 1 - c\langle \mathbf{h}, \mathbf{v}\rangle\}, \tag{8}$$

where $\mathbf{v}_i$ represents the feature vector generated by the $i$-th sample, and $c_i \in \{+1, -1\}$ represents the class label. The notation $\langle \mathbf{h}, \mathbf{v}\rangle$ represents the inner product of vectors **h** and **v**. The passive aggressive algorithm [31] is used to update the hyperplane parameters effectively:

$$\mathbf{h} \leftarrow \mathbf{h} - \frac{\ell\big(\mathbf{h}; (\mathbf{v}, c)\big)}{\big\|\nabla_h \ell\big(\mathbf{h}; (\mathbf{v}, c)\big)\big\|^2 + \frac{1}{2\tau}} \nabla_h \ell\big(\mathbf{h}; (\mathbf{v}, c)\big) \tag{9}$$

where the gradient of the loss function **h** is denoted by $\nabla_{\mathbf{h}} \ell\big(\mathbf{h}; (\mathbf{v}, c)\big)$, and $\tau \in (0, \infty)$ is a hyper-parameter used to control the **h** update rate. Similar to the long-term filter $f_{Long}$, we use Eq. (9) to update the classifiers parameters only when $C_{t,Long} \geq T_a$.

### 3.4 Confidence Function and Update Strategy

The tracking confidence parameter is an important index for judging whether the target is lost. Most CF-based trackers use maximum response $R_{\max}$ to locate the target in the next frame, but in a complex scene, it is not ideal to rely on this parameter alone. Wang et al. proposed LMCF [24] with average peak-to-correlation energy (APCE, Eq. (10)), which can effectively deal with the target occlusion and loss. Zhang et al. proposed a tracker MACF [25] based on confidence of squared response map (CSRM, Eq. (11)), and used to make up for the lack of APCE discrimination during long occlusion:

$$APCE = \frac{|R_{max} - R_{min}|^2}{mean\Big(\sum_{w,h}\big(R_{w,h} - R_{min}\big)^2\Big)}, \tag{10}$$

$$CSRM = \frac{\big|R_{max}^2 - R_{min}^2\big|^2}{mean\Big(\sum_{w,h}\big(R_{w,h}^2 - R_{min}^2\big)^2\Big)}, \tag{11}$$

where $R_{max}$, $R_{min}$, and $R_{w,h}$ respectively denote the maximum, the minimum, and the $w$-th row $h$-th column elements of the peak value of the response respectively. To determine whether the combination of multiple confidence parameters will improve the performance of the tracker, a comparative was conducted, and the details are presented experiment in Section 4.

In the traditional DCF-based tracker [5-7], the general practice is to train a sample and the filter at each frame, then update filter. Although iterative search can be conducted effectively, updating the filter for each frame has a serious impact on the computing load because the optimization calculation of the filter is the core calculation step in the entire algorithm. To improve the computational efficiency, we adopt ECO [8] method to reduce the computational complexity by updating the filter template in every $N_S$-th frame. This can improve the running speed and effectively prevent over-fitting. Similarly, the target sample is updated in each frame.

### 3.5 Solutions in Low-Illumination Environments

The image enhancement algorithm has been addressed in many intriguing works [26-28], and in the present study, the project presented by Guo et al. [26] is used to implement a baseline using the standard LIME. For a low-light image, the model can be expressed in the following two forms:

$$\mathbf{L} = \mathbf{R} \odot \mathbf{T}, \tag{12}$$

$$1 - \mathbf{L} = (1 - \mathbf{R}) \odot \widetilde{\mathbf{T}} + \alpha\left(1 - \widetilde{\mathbf{T}}\right), \tag{13}$$

where $\mathbf{L}$ is the captured image, $\mathbf{R}$ is the desired recovery, $\mathbf{T}$ represents the illumination map, and $\alpha$ is the global atmospheric light. The physical meaning of the first model (Eq. (12)) is that the observed image can be resolved into the product of the desired scene and the illumination image. The second model (Eq. (13)) [29] is based on inverted low-light images $1 - \mathbf{L}$ , which look like haze images. LIME [26] divides image enhancement into initial illumination map estimation and illumination map refinement. In first step, the following preliminary initial estimates of non-uniform lighting for each individual pixel $p$ are used:

$$\widehat{\mathbf{T}}(p) \leftarrow \max_{c \in \{R,G,B\}} \mathbf{L}^c(p), \tag{14}$$

Too ensure that the obtained $\widehat{\mathbf{T}}(p)$ recovery will not be saturated; a small constant $\varepsilon$ is defined to avoid a zero denominator:

$$\mathbf{R}(p) = \mathbf{L}(p) / \left(\max_c \mathbf{L}^c(p) + \varepsilon\right), \tag{15}$$

Afterwards, the atmospheric light α is substituted into the $1 - \mathbf{L}$ model (Eq. (13)):

$$\widetilde{\mathbf{T}}(p) \leftarrow 1 - \alpha^{-1} + \max_c \mathbf{L}^c(p) \cdot \alpha^{-1}, \tag{16}$$

$$\mathbf{R}(p) = \frac{\mathbf{L}(p) - 1 + \alpha}{\left(1 - \alpha^{-1} + \max_c \mathbf{L}^c(p) \cdot \alpha^{-1} + \varepsilon\right)} + (1 - \alpha), \tag{17}$$

We obtained the initial illuminance $\widehat{\mathbf{T}}(p)$ by Eq. (14) due to its conciseness, and the refined illumination map $\mathbf{T}$ can be expressed as the following optimization problem:

$$\min_{\mathbf{T}} \left\|\widehat{\mathbf{T}} - \mathbf{T}\right\|_F^2 + \beta \|\mathbf{W} \odot \nabla \mathbf{T}\|_1, \tag{18}$$

where$\beta$ is a regularization weight, $\|\bullet\|_F$ and $\|\bullet\|_1$ designate the Frobenius and $\ell_1$ norms, respectively. $\mathbf{W}$ is a weight matrix, and $\nabla$ is the first-order derivative filter consisting of $\nabla_h \mathbf{T}$ (horizontal) and $\nabla_v \mathbf{T}$ (vertical). For the selection of weight matrix $\mathbf{W}$, inspiration was drawn from RTV [30], and the two-throughout Gaussian kernel $G_\sigma(p, q)$ was used as the standard deviation $\sigma$. For each location, the weight (e.g. $\mathbf{W}_h(p)$) can be set in the following manner:

$$\mathbf{W}_h(p) \leftarrow \sum_{q \in \Omega(p)} \frac{G_\sigma(p, q)}{\left|\sum_{q \in \Omega(p)} G_\sigma(p, q) \nabla_h \widehat{\mathbf{T}}(q)\right| + \varepsilon}, \tag{19}$$

where $\Omega(p)$ is a region centered at pixel $p$, $q$ is the location index within the region, and $|\bullet|$ is the absolute value operator. Eq. (18) can thus be approximately calculated by the following:

$$\min_{\mathbf{T}} \left\|\widehat{\mathbf{T}} - \mathbf{T}\right\|_F^2 + \beta \sum_p \frac{\mathbf{W}_h(p)\left(\nabla_h \mathbf{T}(p)\right)^2}{\left|\nabla_h \widehat{\mathbf{T}}(p)\right| + \varepsilon} + \frac{\mathbf{W}_v(p)\left(\nabla_v \mathbf{T}(p)\right)^2}{\left|\nabla_v \widehat{\mathbf{T}}(p)\right| + \varepsilon}, \tag{20}$$

Hence, Eq. (14) can be employed to initially estimate the illumination map $\widehat{\mathbf{T}}$. After the refined illumination map $\mathbf{T}$ is obtained by Eq. (18), $\mathbf{R}$ can be recovered by Eq. (15). However, LIME is mainly employed to carry out the enhancement of single picture, without considering the real-time performance of video. Therefore, a simpler image enhancement method is proposed as a baseline.

According to the brightness extraction formula in Max-RGB [28]:

$$Light = (\max_{c\in\{R,G,B\}} \mathbf{L}^c(p) + \min_{c\in\{R,G,B\}} \mathbf{L}^c(p))/\mathbf{2}\,, \tag{21}$$

The average values of the $R$, $G$, and $B$ channels are respectively calculated, and the maximum and minimum mean values are taken as the brightness values. An effective non-linear superposition algorithm is iterated to improve the brightness of the image:

$$\mathbf{R}(p) = \sum\nolimits_{c\in\{R,G,B\}} (\mathbf{L}^c(p) + k\mathbf{L}^c(p) \odot (1 - \mathbf{L}^c(p)/256))\,, \tag{22}$$

where $\mathbf{L}(\bullet)$ denotes the captured image, $\mathbf{R}(\bullet)$ denotes the desired image, and $k \in [0,1]$ is a parameter for controlling exposure. Eq. (22) is iterated until the brightness reaches the threshold value, and the feature of the target is then extracted for filter training.

Because the global brightness cannot effectively represent whether the target is already in a low-illumination environment (Fig. 2), the scope of perception is limited to filled bounding boxes instead of using global brightness awareness to avoid the impact of high-brightness areas on the areas that are actually needed. When the brightness is below the predetermined threshold $T_L$, the exposure of the current frame is compensated, and skip this step when the light is sufficient.

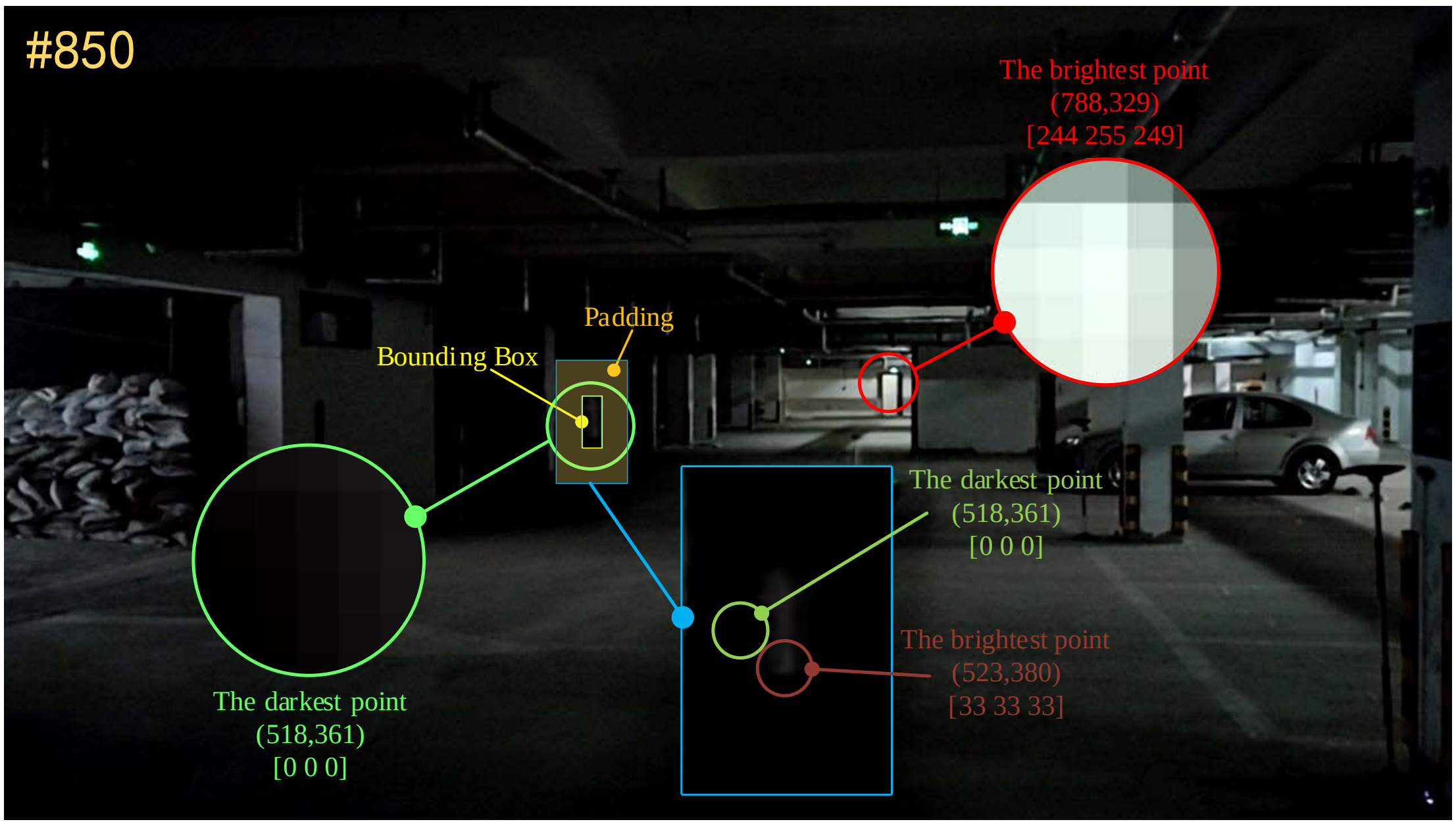

**Fig. 2.** Selection of low-illuminance detection area. In the full view (1280×720), the darkest pixel (green circle) is around the target but the brightest (red circle) is not. The maximum RGB around the target is only [33 33 33]. That is, the global brightness of the image is *Light*=128, but the brightness around the target is only *Light* =16.5.

## 4 Experimental

### 4.1 Implementation Details

We implement our method LRCF in MATLAB 2016b on an Intel i7 3.70 GHz CPU and Nvidia GeForce GTX 1080 GPU. To adapt to the working conditions of dark light, LLCF added only the image enhancement module to the LRCF. The CNN feature was not applied to improve the real-time performance of the operation, and only the hand-craft features (HOG and color names) were used. The dataset adopted to evaluate our method was the Online Tracking Benchmark (OTB) [31], which is an authoritative dataset that was used to compare with other state-of-the-art trackers. It contains 50 sequences with many challenging attributes.

The regularization parameter $\lambda$ was set to 0.01, and the learning rate $\eta$ was set to 0.025. The standard deviation of the Gaussian function output $y$ was set to 1/16 of the translation target size. The padding of the filter was set to two times the size of the initial target. The scale filters were interpolated from $N$ = 17 scales to $N^*$ = 33 scales by interpolation and the scale factor of $a$ = 1.02. The re-detection threshold $T_r$ was set as 0.2 for the activation detection module, and $T_a$ = 0.4 for the detection result. Note that the threshold setting here is only the corresponding fraction of the long-term filter $\mathbf{f}_{Long}$. The confidence parameters of the response of LCT were respectively set, as 0.9 times the maximum response and 0.75 times the APCE (or CSRM). The long-term memory template was updated when both of its parameters exceeded the set threshold. The purpose of this was to test the effect of multi-confidence settings on the performance of trackers. On LLCF, the exposure control parameter $k$ was set to 1. The lighting intensity threshold $T_L$ = 48. The number of iterations for image enhancement was 1 (Section 4.4).

The OTB dataset [31] results were evaluated by overlapping precision (OP), distance precision (DP), and tracking speed (FPS). The success plots were charted in terms of the distance precision rate [0,1] in 20 pixels and the area-under-the-curve (AUC) of the overlap success rate.

### 4.2 Update Strategy Comparison

The update gap was set as $N_S$ = 1, 3, 5, and the performance (AUC) was compared with the running speed. The experimental results are presented in Table 1 and Fig. 3. It was found that it was best to update the filter when $N_S$ = 3, but that the running speed was significantly improved when $N_S$ = 5.

**Table 1.** Update strategy comparison

| $N_S$ | AUC | FPS |
| --- | --- | --- |
| 1(per frame) | 58.6 | 29.5 |
| 3 | 61.3 | 32.8 |
| 5 | 60.6 | 37.5 |
| Fast DSST | 60.0 | 173 |

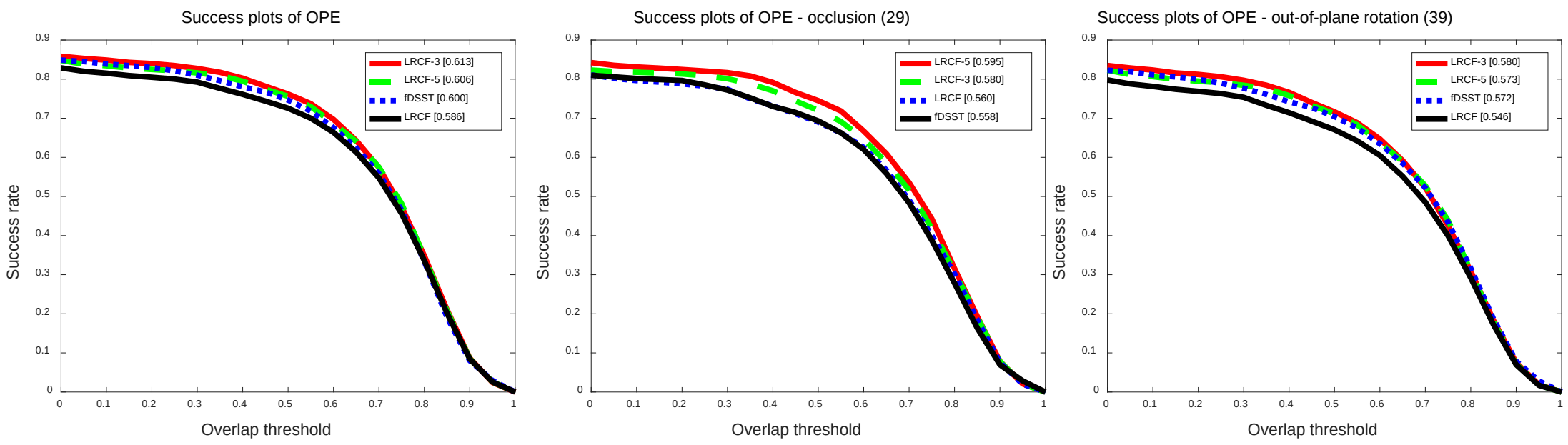

**Fig. 3.** The OTB-2013 benchmark test for the interval update strategy. The effect was obviously improved after adding long-term memory filter. In an occlusion environment, the performance when $N_S = 3$ was not as good as when $N_S = 5$, presumably because of the occlusion time, and the occlusion object is trained as a sample.

### 4.3 Overall performance

The proposed method was compared with the trackers presented in the work by Wang et al. [31], and other trackers including the baseline trackers TLD [18], LCT [19], and fast DSST [22]. Fig. 4 presents the comparison of our method with the baseline trackers. Via comparison, it can be determined that our method is superior; it has an average running speed of 33 FPS ($N_S = 3$), and can run in real time. Fig. 5 demonstrates the superiority of the proposed method in difficult tracking scenarios, such as low resolution, fast motion, scale variation, and occlusion. Compared with fast DSST, the proposed method presents improvements in distance precision of 12.3%, 2.0%, 1.4%, and 6.6%, respectively. It was also revealed that using multiple confidence parameters in filter updates can lead to performance degradation.

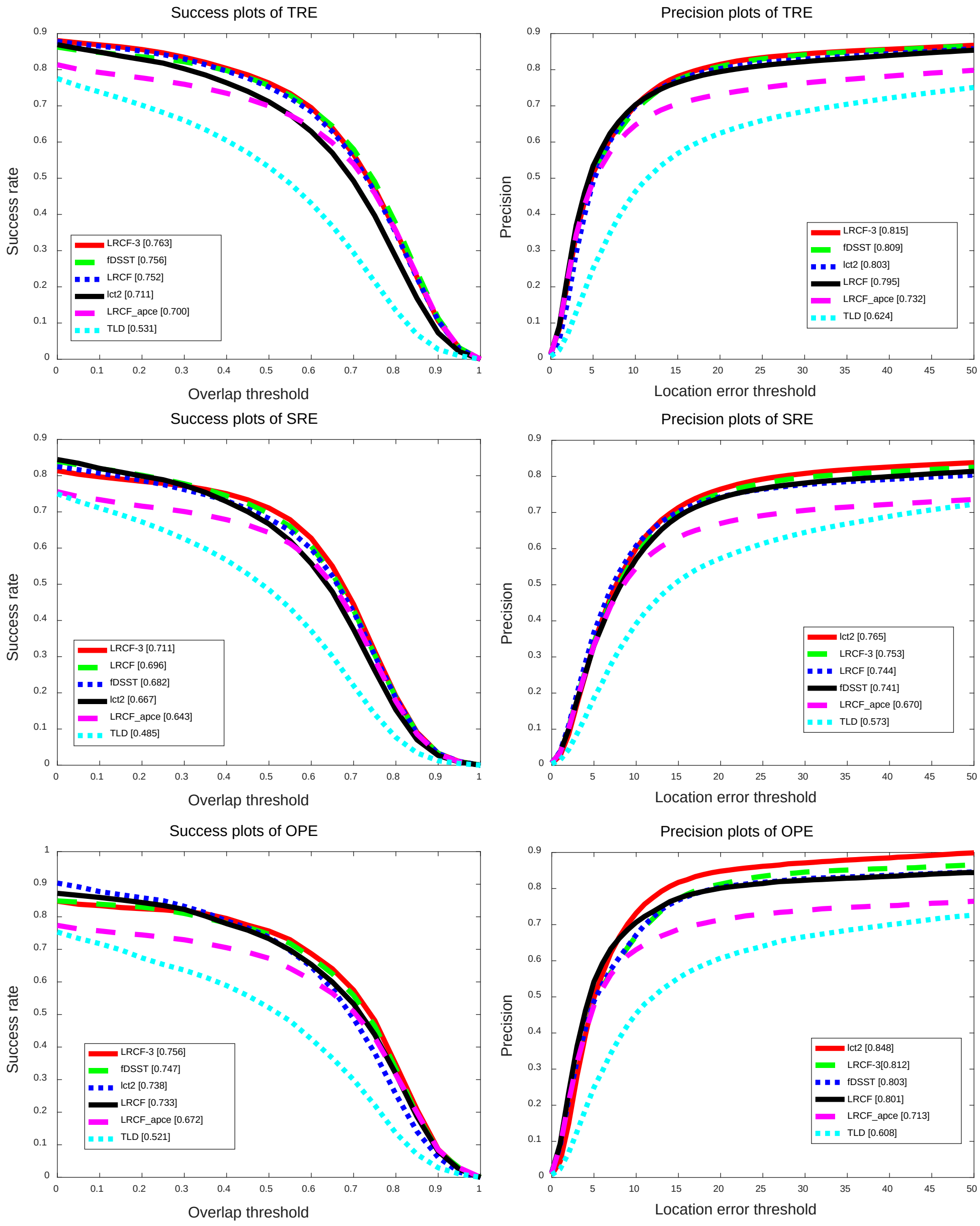

**Fig.4.** Distance precision and overlap success plots of the average overall performance on the OTB dataset. One pass evaluation (OPE) and temporal robustness evaluation (TRE) are presented in this figure. OPE refers to the success threshold of calculating the proportion of successful tracking frames in the total video frames in the evaluation dataset. TRE refers to the success threshold of calculating the proportion of successful tracking frames in the total video frames after 20 different sampling areas (deviation around the target).

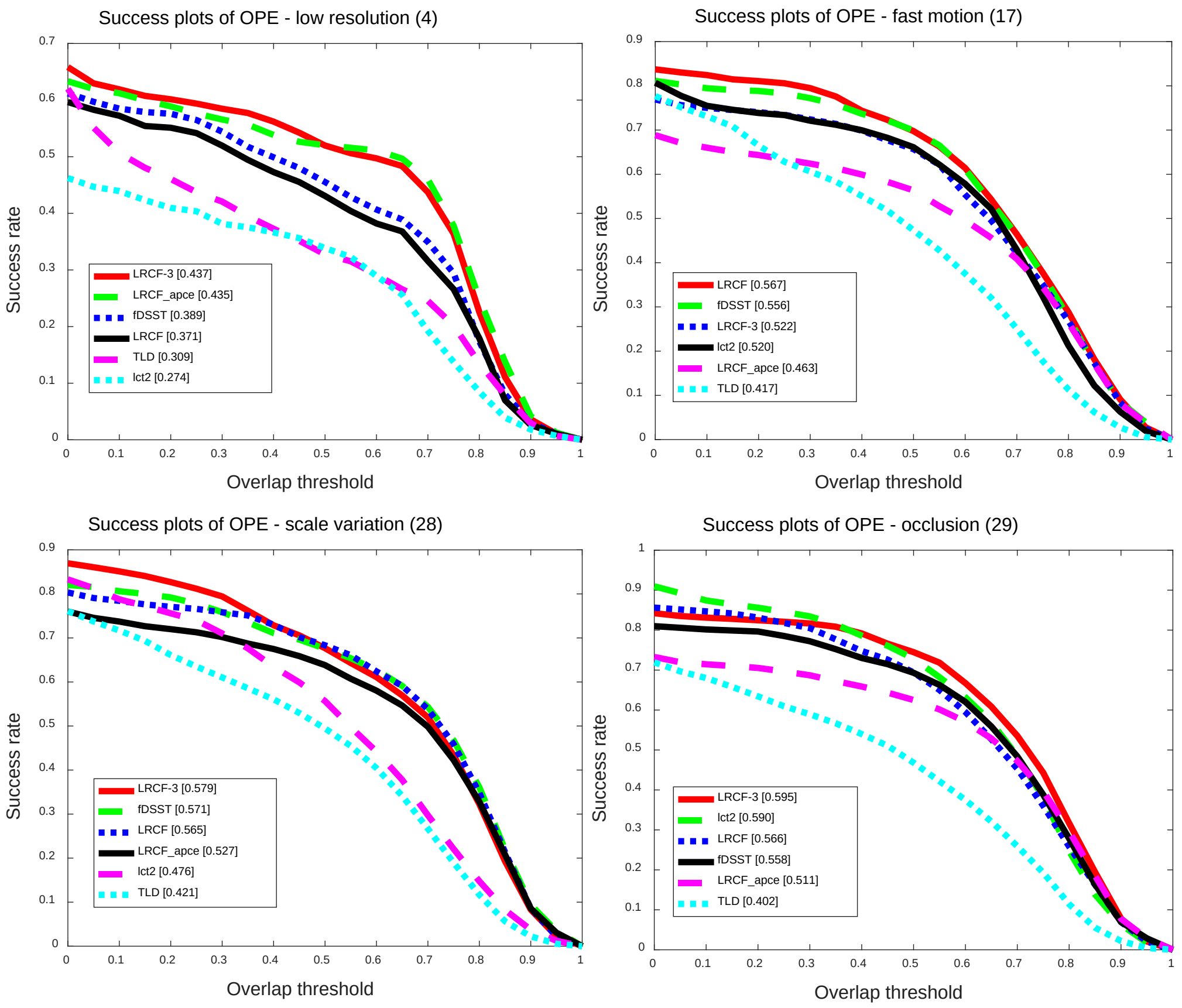

**Fig.5.** The comparison results of the proposed method and the baseline under different types of challenges. The area-under-the-curve (AUC) score for each tracker is reported in the legend.

### 4.4 Experiments of Low-Illumination Environments

To detect the performance of the algorithm under low-illumination conditions, we collected the sequence in an underground garage. The video is 34 seconds long and comprises, a total of 1044 frames. The target was subject to light change, out-of-plane rotation, background similarity, and an almost completely dark area from the 800$^{th}$ frame.[1]

The effect of illumination discrimination on image enhancement is shown in Fig. 6. On the *David* sequence, a bright spot can be seen in the lower left corner of the image in the 83$^{rd}$ frame, which causes the global brightness to be as high as 120, although the brightness around the actual target is only 21. The bounding box (BB) discrimination performs well in determining enhancement success, as the global illumination is 120 on the 83$^{rd}$ frame. The standard *David* sequence is calibrated from the 300$^{th}$ frame, but the target is already visible at the 150$^{th}$ frame ($Light = 30$). From this sequence alone, it does not seem like the threshold $T_L$ needs to be set very high, but in the *Garage* sequence the disadvantage of this situation is evident (Fig. 2). In Fig. 6, which depicts the *Garage* sequence after 800$^{th}$ frame, the global illuminance is maximum, but the BB illuminance is almost less than 20. The target cannot be distinguished from the surrounding environment. The exposure control parameter *k* was set to 1, and the lighting intensity threshold $T_L$ was conservatively set as 48.

[1] The sequence is available at https://drive.google.com/open?id=1WQC8-iHxxpV5jjrmaZVTOo5u0IZ7thaH, and the demo video has been uploaded on YouTube at https://youtu.be/m9AwamHZ4uY.

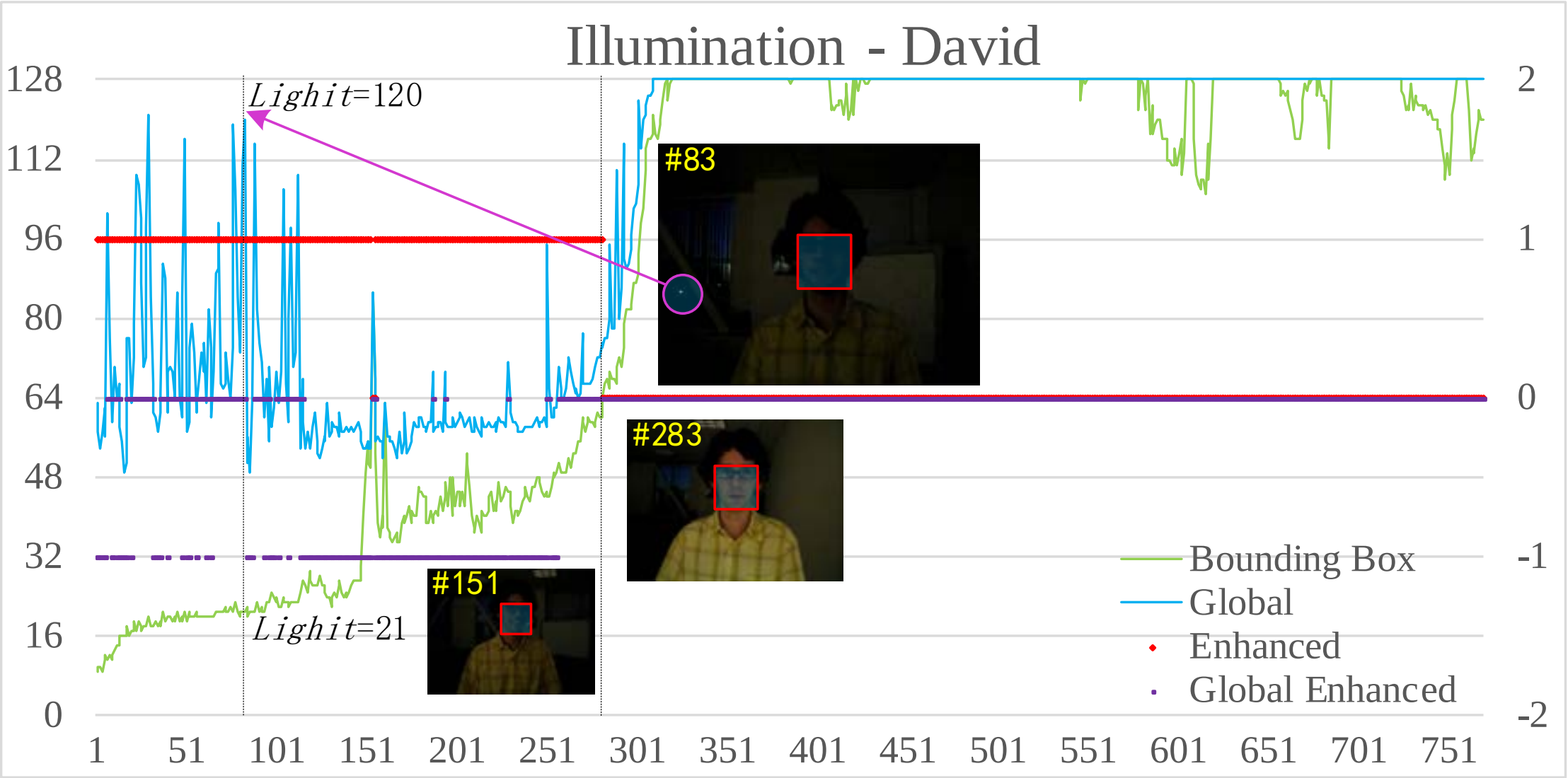

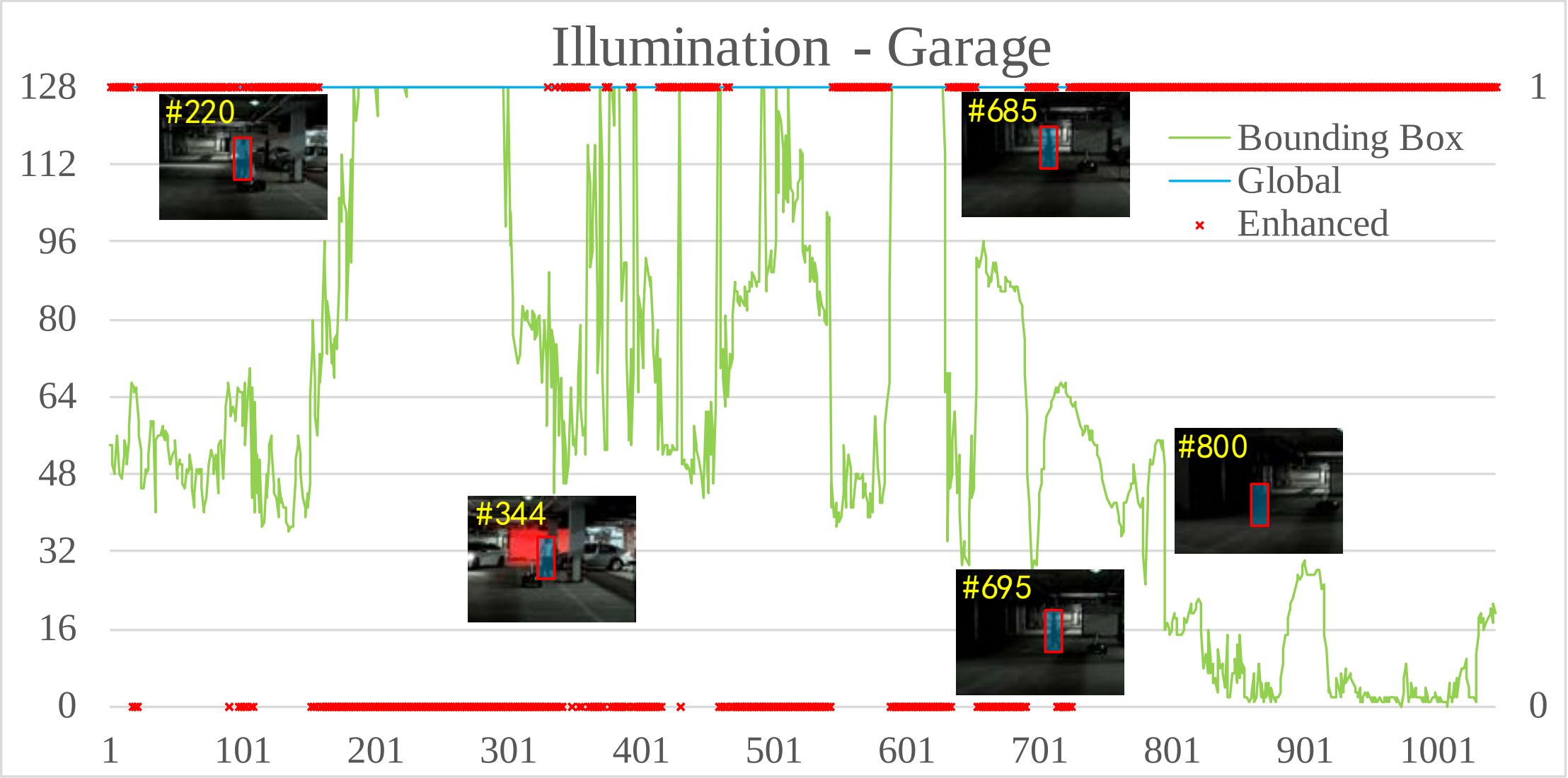

**Fig.6.** Comparison of global and bounding box (BB) illumination. The above figure shows the luminance value distribution of the *David* sequence, and the following is the *Garage* sequence. It can be seen that both BB and global luminance increase with time in the *David* sequence. However, in the *Garage* sequence, the global brightness is always the maximum because of the lamp on the right.

Fig. 7 presents the comparison of the two modes of LIME [26], MATLAB function *imadjust*, and our method in terms of time cost. The first picture shown in Fig. 7 presents the comparison curves, of the four kinds of solvers (Ours, *imadjust*, LIME-exact, and LIME-sped-up solvers) in terms of time cost. The sample pictures are scaled to different levels from the original size (3144×3078). From the curves, it is evident that when the image size is smaller than 400, the four solvers are sufficiently efficient. However, it can be seen from the histogram, that when the image size is small, the LIME solvers (the exact solver and the sped-up solver) take on the order of seconds to calculate a frame, our method takes only on the order of milliseconds to perform an iterative calculation. For a 720P resolution image, our method is 20 times faster than LIME-sped-up. And in the same level of comparison, our method slightly faster than the MATLAB function. After the image size exceeds 1000, there is a sharp difference in the time cost of the four enhancement methods. The frame rate of our method is still on the order of milliseconds, and LIME has entered the time cost on the order of seconds.

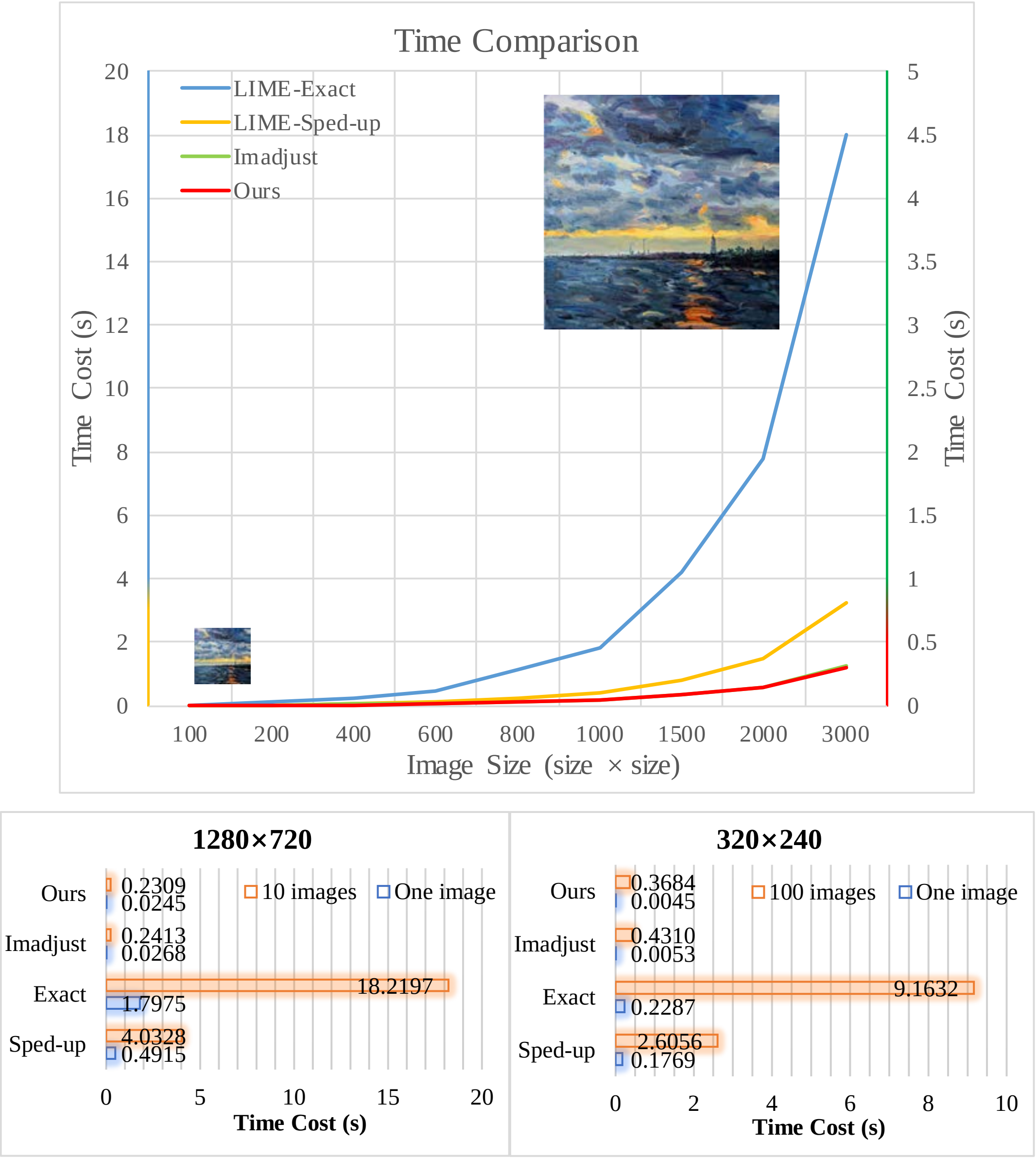

**Fig. 7.** Comparison of the cost of several image enhancement methods with varying image sizes. **Top**: The left vertical scale indicates LIME solvers, and the right indicates the other two solvers. **Left**: Input image resolution 1280×720. **Right**: Input image resolution 320×240.

Fig. 8 presents the comparison of our tracker (yellow) with the benchmark trackers, e.g. LRCF (blue), DSST (green) and LCT (red). The proposed tracker was also compared with the state-of-the-art trackers BACF [7] (orange) and ECO [8] (cyan). In the 165th frame (background similarity), the LCT and ECO fail, following the car after the 320th frame. After the 810th frame into the dark area, LRCF, DSST and BACF remain in their original positions, and only LLCT tracks to the last frame.

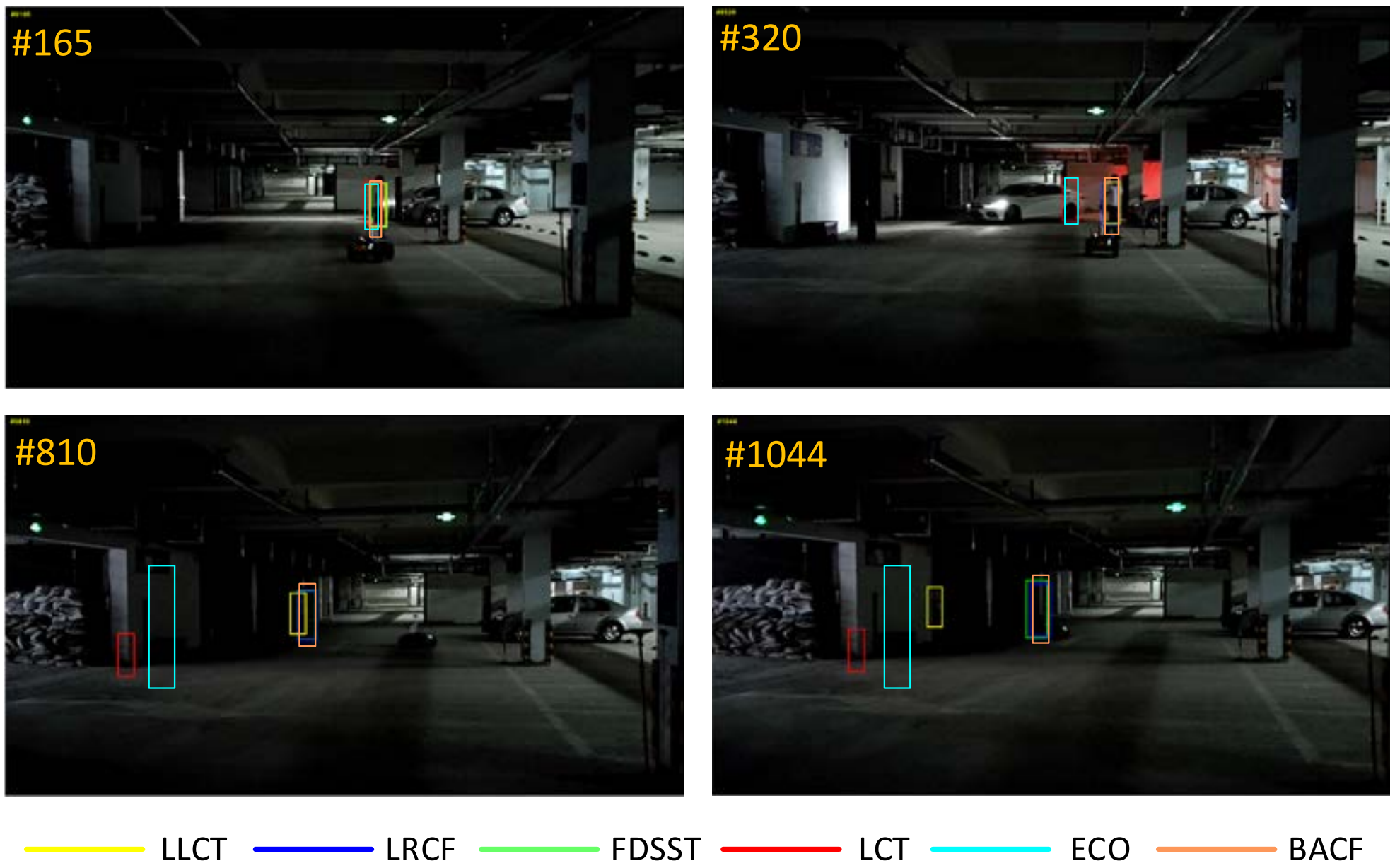

**Fig.8.** Field experiment in the underground garage. The proposed method performed well in a low-illumination environment.

Fig. 9 shows that the reproduction of image details in LIME [26] is excellent, the noise control is effective, and only the overexposed area appears on the left side of the image. Relatively speaking, the "rude" method is not excellent in image noise reduction control and almost all areas other than the low-illumination area are overexposed. In practical application, LIME better addresses the restoration of image details. Via experimentation (Figs. 8 and 9), it was determined that for target tracking, when the model is hidden in the dark, the effect of the accurate enhancement method is almost the same as that of the "rude" enhancement method. Additionally, the rude method has a smaller burden, and hardly affects the real-time performance of the tracker. This means that the tracker does not need "hyperfine" sample model to be followed correctly, and it can also track the target correctly according to the difference between the target and the background.

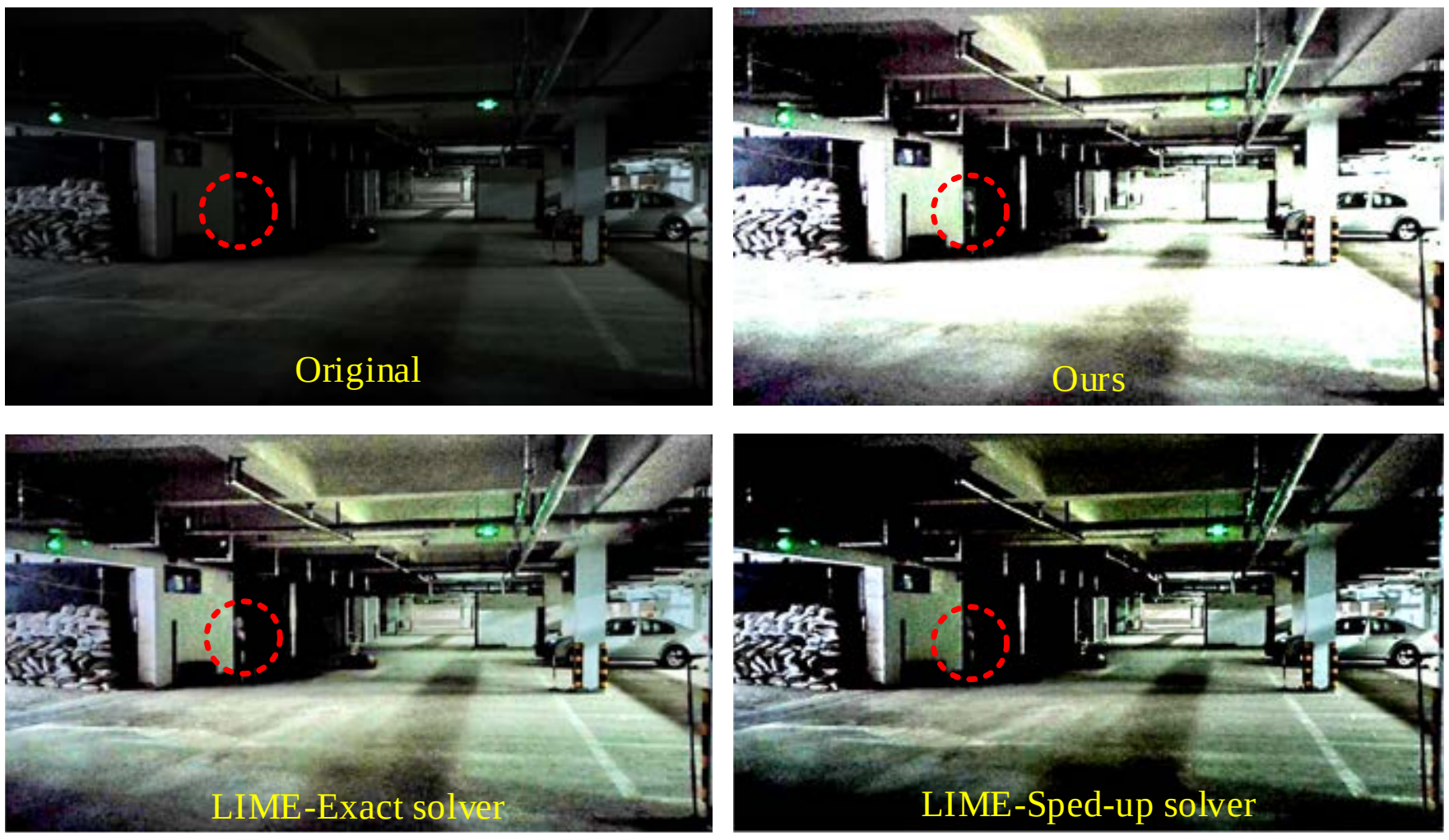

**Fig.9.** A professional image enhancement algorithm, LIME [26], was used to restore the position of the last frame (1044th) of the target, and was compared with the proposed method.

## 5 Conclusions

In this paper, an effective tracker, LLCT, was proposed for long-term, low-illumination visual tracking based on the fast DSST framework. First, LRCF was learned, a projection matrix was introduced to the traditional DCF filter, and the dimension of the extracted sample features was reduced. Then, the concept of fast DSST was used to calculate the correlation of images instead of kernel convolution. The experimental results demonstrated that LRCF has good robustness and effectiveness for the OTB-50 benchmark. To achieve the tracking task in low-illumination environments, an image enhancement module was added and an ablation study was conducted. Field experiments revealed that the proposed LLCT is superior to other methods in low-illumination environments. However, there still exist many shortcomings; for example, for a large size image, the enhancement module takes up much of the computational cost, and is thus a research direction that must be taken in the future.

## Acknowledgments

This work has been supported by grant of the National Key Research and Development Program of China (No. 2018YFC0808000); the Priority Academic Program Development of Jiangsu Higher Education Institutions of China (PAPD); the Natural Science Foundation of the Jiangsu Higher Education Institutions of China (No.19KJB460014).